\PassOptionsToPackage{dvipsnames}{xcolor}

\documentclass{article}
\usepackage{bbm}
\usepackage[utf8]{inputenc} 
\usepackage[T1]{fontenc}    
\usepackage{hyperref}       
\usepackage{url}            
\usepackage{booktabs}       
\usepackage{amsfonts}       
\usepackage{nicefrac}       
\usepackage{microtype}      
\usepackage{lipsum}		
\usepackage{graphicx}
\usepackage{float}
\usepackage{natbib}
\usepackage{doi}
\usepackage{xcolor}
\usepackage[normalem]{ulem}

\usepackage{color}
\usepackage{lineno}
\usepackage{amsmath}
\usepackage{xcolor}

\usepackage{soul}
\usepackage{arxiv}
\usepackage{url}
\usepackage{bm}
\newcommand\myshade{85}
\colorlet{mylinkcolor}{RoyalBlue}
\colorlet{mycitecolor}{violet}
\colorlet{myurlcolor}{black}

\hypersetup{
  linkcolor  = mylinkcolor!\myshade!black,
  citecolor  = mycitecolor!\myshade!black,
  urlcolor   = myurlcolor!\myshade!black,
  colorlinks = true,
}

\usepackage{times}
\usepackage{latexsym}

\usepackage[T1]{fontenc}

\usepackage[utf8]{inputenc}

\usepackage{microtype}

\usepackage{booktabs}
\usepackage{graphicx}
\usepackage{array}
\usepackage{multirow}
\usepackage{longtable}
\usepackage{CJKutf8}
\usepackage{float}
\usepackage{microtype}
\usepackage{multirow}
\usepackage{booktabs}
\usepackage{graphicx}
  
\usepackage{amssymb}
\usepackage{url}
\usepackage{hyperref}
\usepackage{amsmath}
\usepackage{caption}
\usepackage[ruled,vlined,linesnumbered]{algorithm2e}
\usepackage{subfigure}
\usepackage{makecell}
\usepackage{ulem}
\usepackage{algorithmic}
\usepackage[misc]{ifsym}
\usepackage{lipsum}
\usepackage{colortbl}

\usepackage{tikz}
\usetikzlibrary{shapes.geometric}

\usepackage{palatino} 

\title{Deep Reinforcement Learning with Task-Adaptive Retrieval via Hypernetwork}

\newcommand\blfootnote[1]{%
\begingroup
\renewcommand\thefootnote{}\footnote{#1}%
\addtocounter{footnote}{-1}%
\endgroup
}

\author{ 
  \small
    Yonggang Jin \textsuperscript{1},\space 
    Chenxu Wang\textsuperscript{1},\space
    Tianyu Zheng\textsuperscript{1},\space \\
    \textbf{Liuyu Xiang} \textsuperscript{1},\space
    \textbf{Yaodong Yang}  \textsuperscript{3},\space
    \textbf{Junge Zhang} \textsuperscript{4},\space
    \textbf{Jie Fu} \textsuperscript{2}$^{*}$,\space
    \textbf{Zhaofeng He} \textsuperscript{1}$^{*}$\\    
{\small 
\textsuperscript{1} Beijing University of Posts and Telecommunications;
}\vspace{-0.1mm} \\ 
{\small 
\textsuperscript{2} Beijing Academy of Artificial Intelligence;
}\vspace{-0.1mm} 
{\small 
\textsuperscript{3} Peking University;
}\vspace{-0.1mm} 
{\small 
\textsuperscript{4} CASIA;
}\vspace{-0.1mm}\\
}

\begin{document}

\maketitle

\begin{abstract}
Deep reinforcement learning algorithms often suffer from sampling inefficiency, as they require numerous interactions with the environment to develop precise decision-making skills. 
Conversely, humans use their hippocampus to recall relevant information from past experiences, aiding decision-making in new tasks without solely relying on environmental interactions. 
However, creating a hippocampus-analogous module for an agent to integrate past experiences into existing reinforcement learning algorithms poses significant challenges.
\textbf{The first challenge} is the selection of pertinent past experiences for the task at hand, while \textbf{the second challenge} involves their incorporation into the decision-making network.
To overcome these obstacles, we introduce a novel method  utilizes a retrieval network based on task-conditioned hypernetwork, which adapts the retrieval network's parameters depending on the task. 
At the same time, a dynamic modification mechanism enhances the collaborative efforts between the retrieval and decision networks. 
We evaluate the proposed method across various tasks within multitask scenarios in the Minigrid environment.
The experimental results demonstrate that our proposed method significantly outperforms strong baselines. 
Our code and data are available at {\hypersetup{urlcolor=YellowOrange}\url{https://github.com/ygjin11/task-hypernet}}.
\end{abstract}

\blfootnote{* Corresponding Author}
\blfootnote{\href{mailto:daze@bupt.edu.cn}{daze@bupt.edu.cn}, \href{mailto:jie.fu@polymtl.ca}{jie.fu@polymtl.ca}, and \href{mailto:zhaofenghe@bupt.edu.cn}{zhaofenghe@bupt.edu.cn}}

\section{Introduction}
Deep reinforcement learning algorithms have exhibited superior performance in various fields such as gaming ~\citep{schulman2015}, robotic control, and manipulation ~\citep{levine2016, lillicrap2015}. 
However, they suffer from significant data inefficiency. 
Deep reinforcement learning systems necessitate tens of millions of interactions with a game emulator, a process vastly different from human skill acquisition ~\citep{mnih2015}. 
Unlike reinforcement learning algorithms, humans do not depend solely on interactions with their environment to learn new skills.
Instead, they utilize past experiences, stored in the hippocampus, to inform their decision-making processes ~\citep{mcclelland1996, lengyel2007}.
The hippocampus facilitates the connection between the brain's decision-making system and long-term memory by storing and retrieving past experiences from different brain regions ~\citep{mcnaughton1987, eichenbaum2004}.
Furthermore, hippocampal activation is modulated by the nature of the activity, which is reflected in task-specific neural codes  ~\citep{davachi2003}. 
This suggests that modeling hippocampal function could address the data inefficiency challenge in reinforcement learning by introducing an adaptive memory retrieval module to support the learning agent.

\begin{figure}[!htbp]
\centering 
\includegraphics[width=0.80\linewidth]{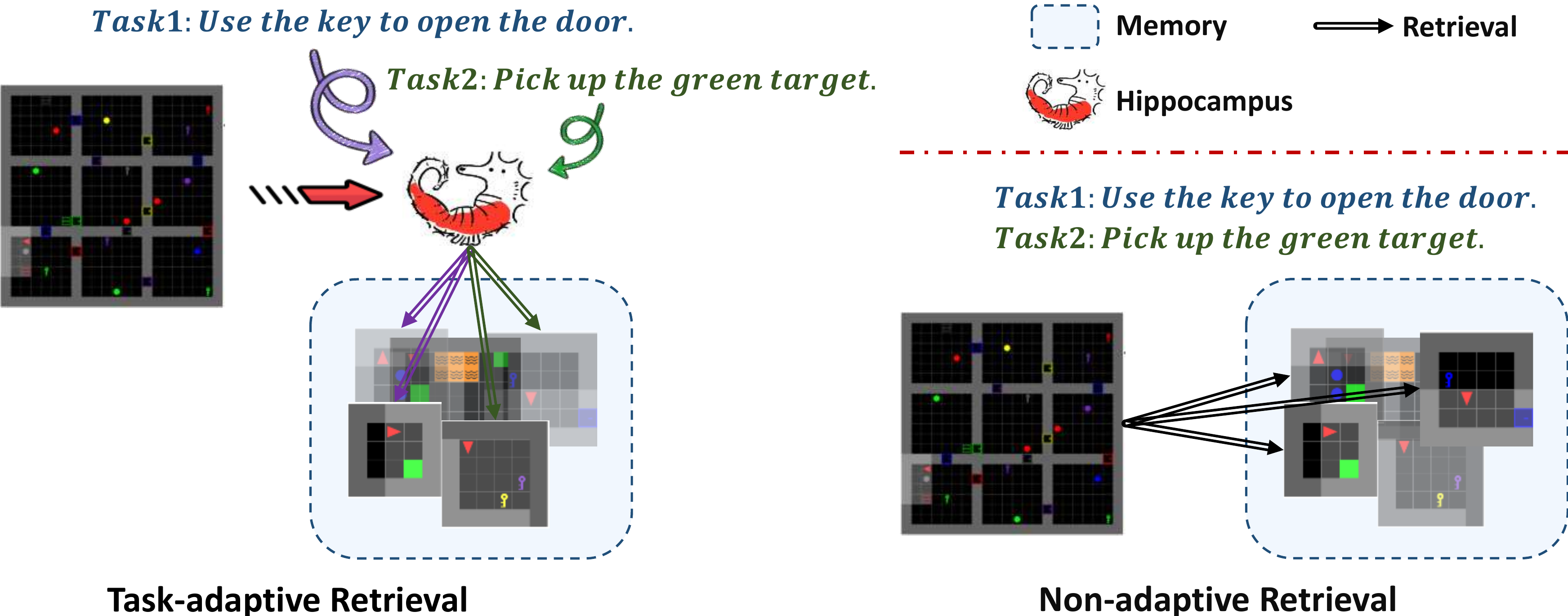}
\caption{On the left, we present an illustrative example of \textbf{Task-Adaptive Retrieval}. Here, the hippocampus of the agent must adapt its retrieval strategies to accommodate the differing demands of Task 1 and Task 2. It retrieves different information for different tasks. In contrast, \textbf{Non-Adaptive Retrieval}, depicted on the right, relies solely on the observed state, leading to consistent outcomes across both tasks due to the lack of task-specific adaptability.} 
\label{fig_introduction}
\end{figure}

~\citet{blundell2016} and ~\citet{pritzel2017} leverage episodic memory to store crucial past experiences, presenting a viable approach. 
Drawing on this foundation, recent works, including ~\citep{lin2018, hansen2018, kuznetsov2021, chen2022, hu2021, goyal2022}, explore the synergy between the rapid convergence facilitated by episodic control and the robust generalization capabilities of reinforcement learning algorithms. 
This is achieved by either integrating episodic memory data into a parametric model or enhancing a parametric model with episodic memory features.  
~\citet{goyal2022} introduces R2A, a model that employs a parametric neural network to retrieve information from trajectories across various tasks, underscoring the benefits of leveraging multitask experiences.
Although previous works on episodic memory create a hippocampal module for agents, their retrieval mechanisms were often constrained in adapting to diverse tasks.
In contrast, human hippocampal retrieval mechanisms are known to be task-dependent ~\citep{davachi2003}.
Similarly, it is essential for agents to have the ability to modify their retrieval mechanisms to suit different tasks.
Figure \ref{fig_introduction} offers an illustrative instance of this phenomenon.
\vspace{0.2cm}

\vspace{-0.15cm}
This study seeks to explore an effective methodology for incorporating a memory retrieval module into reinforcement learning algorithms, aiming to resolve two primary challenges: (1) how to provide the agent with a retrieval network capable of accessing the most pertinent past experiences for a given task, and (2) how to  improve the synergy between retrieval and decision-making networks. 
We propose a novel method that differs from existing algorithms that combine episodic memory with reinforcement learning ~\citep{lin2018, hansen2018, kuznetsov2021, chen2022, hu2021, goyal2022} by adapting the retrieval network to suit different tasks. 
Multitask learning ~\citep{caruana1997} is a fundamental problem in machine learning where the goal is to develop a unified model that performs well on multiple tasks. 
Recently, several works ~\citep{mahabadi2021, ustun2022, kang2023, beck2023} propose learning parameters for different tasks using a shared hypernetwork that conditions on task or module IDs, thereby promoting information sharing across tasks.
Drawing inspiration from these advancements, our method utilizes a hypernetwork to dynamically adapt the parameters of the retrieval network according to the task at hand, thereby addressing the first challenge.
At the same time, A dynamic modification mechanism enhances the collaborative efforts between the retrieval and decision networks to address the second issue. 
Our framework of choice is the stable and efficient Actor-Critic algorithm, Proximal Policy Optimization (PPO)~\citep{schulman2017}, which we augment with a hypernetwork-based retrieval system. 
We evaluate the proposed method across diverse tasks within multitask settings in the Minigrid environment.
Our experimental results demonstrate that the hypernetwork-based retrieval module retrieves task-relevant information for the specific task and significantly improves the performance of PPO. Our contributions are summarized as follows:
\begin{enumerate}
    \item We propose a retrieval network based on task-conditioned hypernetwork, which can retrieve relevant information from episodic memory for the specific task.
    \item We employ a dynamic modification mechanism to regulate the impact of the retrieval network on the decision network, thereby promoting effective collaboration between the two networks.
    \item The retrieval network based on task-conditioned hypernetwork and the dynamic modification mechanism enhance sample efficiency and performance for PPO.
\end{enumerate}

\section{Related Work} 
\vspace{-0.15cm}
\subsection{Episodic Memory} 
\vspace{-0.15cm}
Episodic control involves capturing and retaining the most influential experiences during the learning process to guide an agent's decision-making.
Model Free Episodic Control (MFEC) ~\citep{blundell2016} utilizes episodic memory to store empirical data gathered from interacting with the environment. MFEC retrieves multiple state-action pairs from the episodic memory to predict the optimal action during decision-making.
Neural Episodic Control (NEC) ~\citep{pritzel2017} proposes a semi-tabular differential memory to store past experiences. 
However, both MFEC and NEC suffer from poor generalization. 
This limitation can be addressed by combining episodic memory with reinforcement learning via the distillation of data into a parametric model or by modifying a parametric model with episodic memory.  
Previous works that adopt this process include ~\citep{lin2018, hansen2018, kuznetsov2021, chen2022, hu2021}. 
For instance, ~\citet{lin2018} proposes Episodic Memory Deep Q-network (EMDQN), which employs episodic memory to supervise an agent's training, while ~\citet{kuznetsov2021} proposes Episodic Memory Actor-Critic (EMAC), which modifies the critic network by leveraging episodic memory experience. 
More recently, ~\citet{goyal2022} incorporates a parameterized network to retrieve information from the trajectories of multiple tasks and shape the predictions of deep Q-network in a fully differentiable manner. 
Although previous works using episodic memory attempt to provide a hippocampal module for the agent, their retrieval mechanisms are constrained as they cannot adapt to the task at hand. 
However, the retrieval mechanism of the hippocampus in humans varies according to different tasks ~\citep{davachi2003}. 
In this paper, we attempt to design an adaptable memory retrieval module that aligns with the human hippocampus's characteristics to improve the agent's decision-making capabilities.
\vspace{-0.15cm}
\subsection{Task-Conditioned Hypernetwork} 
\vspace{-0.15cm}
A hypernetwork is used to generates the primary network's weights.
In the context of learning tasks in sequence, ~\citet{von2019} propose a task-conditioned hypernetwork responsible for computing the weights of the target model. 
Recently, some works propose using task-conditioned hypernetwork to generate the parameters of specific network modules, such as the adapter inserted into a transformer layer ~\citep{mahabadi2021,ustun2022}, the decision network connected to the state encoder ~\citep{beck2023}, and the curriculum module ~\citep{kang2023}. 
Among these works, the most relevant to our study is ~\citet{beck2023}, which utilizes a hypernet to compute the parameters of the policy network.
The utilization of hypernetwork improves the network parameter's adaptability to various tasks toward resolving multitask scenarios. 
Our study aims to adapt the retrieval network parameters to each task. We utilize a task-conditioned hypernetwork to provide the memory retrieval module with diverse parameters based on the given task, resembling the cortex's varying activation degrees of the hippocampus when doing different tasks.
\vspace{-0.25cm}
\section{Method}
\vspace{-0.25cm}
\begin{figure}[!htbp]
\centering
\includegraphics[width=0.90\linewidth]{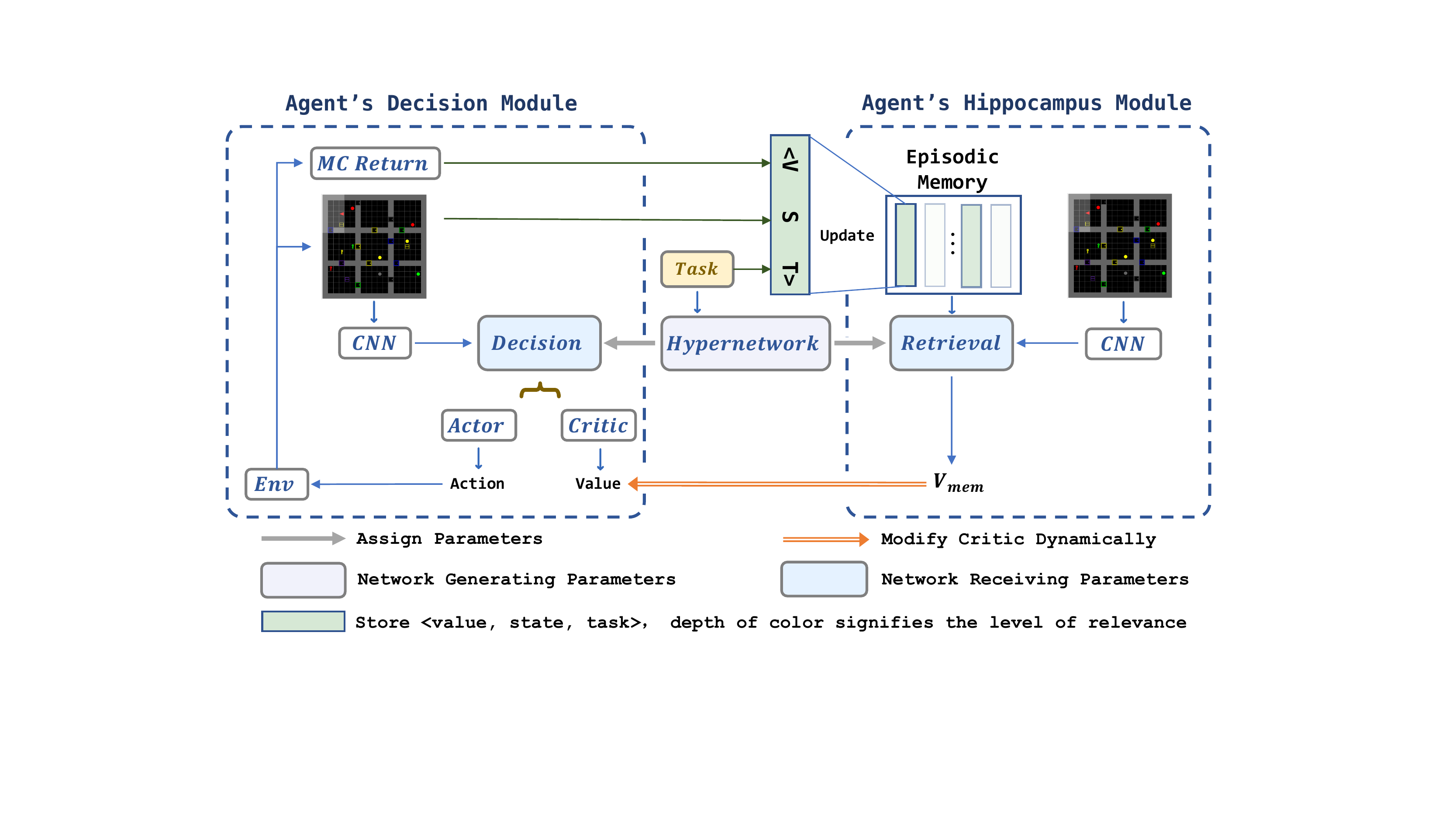}
\caption{The model architecture.  
To start, a task-conditioned hypernetwork dynamically adapts the parameters of both the retrieval and decision networks according to the ongoing task. 
Subsequently, the retrieval network accesses the episodic memory to compute $\bm{V_{mem}}$.
Lastly, the loss of the critic network undergoes modification through the incorporation of $\bm{V_{mem}}$, facilitated by a dynamic modification mechanism. 
As a result, gradients are computed, propagated, and utilized to update the network parameters.} 
\label{fig_method}
\end{figure}

We present a novel method named \textbf{Actor-Critic with Multitask Episodic Memory Based on Hypernetwork}. 
Our approach employs a hypernetwork to adapt the retrieval network’s parameters depending on the task, as discussed in section \ref{rboh}. Additionally, we integrate a dynamic modification mechanism to promote collaboration between the parameterized retrieval network and the decision network,elaborated in section \ref{m_dmm}.
Our approach is developed within the framework of Proximal Policy Optimization (PPO) \citep{schulman2017}. 
We depict our method in Figure \ref{fig_method}. 
The notations utilized in this paper are listed in Appendix \ref{table_notation}.
\vspace{-0.15cm}
\subsection{Retrieval Based on Hypernetwork}
\vspace{-0.15cm}
\label{rboh}
In this section, we delineate the comprehensive process for retrieving the most relevant information for various tasks.  Section \ref{em} elucidates the structure of episodic memory. Section \ref{tch} delineates the process by which the hypernetwork adjusts parameters based on task information. Section \ref{rn} expounds on the process of retrieving pertinent information from past experiences.

\vspace{-0.4cm}
\subsubsection{Episodic Memory}
\vspace{-0.2cm}
\label{em}
Episodic Memory is characterized as a memory table containing triples in the form of $<v_i^{mem}, s_i^{mem}, t_i^{mem}>$. 
Here, $s_i^{mem}$ signifies an environmental state, $t_i^{mem}$ denotes a specific task, and $v_i^{mem}$ corresponds to the Monte Carlo (MC) returns derived upon the conclusion of an episode. 
The memory table is regularly updated with state observations, current tasks, and MC returns, which can only be computed at the conclusion of an episode.

\vspace{-0.2cm}
\subsubsection{Task-Conditioned Hypernetwork}
\vspace{-0.2cm}
\label{tch}
Our objective is to augment the adaptability of the retrieval network across diverse tasks.
The notion of utilizing a hypernetwork to generate network module parameters dynamically has attracted substantial interest in recent times ~\citep{von2019,mahabadi2021,ustun2022,kang2023}.
Our method entails dynamically adapting the parameters of the retrieval network based on the specific task. 
Concurrently, we conceptualize the hypernetwork as a pivotal cognitive center of the agent. It generates parameters not solely for the retrieval network but also for the decision network.
This collaborative parameter generation aims to enhance the collaboration between the decision and retrieval networks.

Two approaches are employed to represent tasks. 
The first approach utilizes one-hot encoding to represent the task ID, which involves creating a vector with a single element set to 1 while the remaining elements are set to 0. 
The second approach involves describing the task using natural language, such as the example "picking up the key and opening the door." 
One-hot encoding effectively distinguishes between diverse tasks, while language embeddings offer insights into task similarities.
In reinforcement learning environments, natural language guidance is provided, as seen in Minigrid ~\citep{minigrid} and BabyAI ~\citep{chevalier2018}.
\begin{equation}
    T = \mathbf{Concat}(\mathbf{Embed}_{\mathbf{id}}(t_{\text{id}}) ,\mathbf{Emb}_{\mathbf{lang}}(t_{\text{lang}}))
\end{equation}
Here, $\mathbf{Embed}_{\mathbf{id}}$ refers to a Multi-Layer Perceptron network tasked with embedding the task ID represented by $t_{id}$. 
On the other hand, $\mathbf{Embed}_{\mathbf{lang}}$ corresponds to the encoder of a pre-trained text-to-text Transformer model ~\citep{lewis2019,raffel2020}, which is responsible for embedding the natural language description $t_{lang}$. We select the T5 Transformer model for $\mathbf{Embed}_{\mathbf{lang}}$.

Hypernetwork generates parameters for both the retrieval and decision network. The design of this hypernetwork follows the structure of the standard hypernetwork model introduced by ~\citep{ha2016}. The generated parameters are obtained through the utilization of the following equations:
\begin{equation}
\label{Retriever_w}
{W_{R}}, {b_{R}}, {W_{D}}, {b_{D}} \mathbf{=} \mathbf{Reshape}(\mathbf{Hypernet}(T))
\end{equation}
\begin{equation}
\label{Retriever_R}
\mathbf{Linear_{R}}, \mathbf{Linear_{D}} \mathbf{=} {W_{R}}, {b_{R}}, {W_{D}}, {b_{D}}
\end{equation}
\begin{equation}
\label{Retriever}
\mathbf{Retriever} \mathbf{=} \mathbf{Linear_{R}} \rightarrow \mathbf{ReLU}
\end{equation}
\begin{equation}
\mathbf{Decision} \mathbf{=} \mathbf{Linear_{D}} \rightarrow \mathbf{ReLU}
\label{Decision}
\end{equation}
Here, $\mathbf{Hypernet}$ consists of a two-layer neural network, and ${W_R}$ as well as ${W_D}$ represent the weight matrices for the retrieval and decision networks. 
Correspondingly, ${b_R}$ and ${b_D}$ symbolize the biases associated with the retrieval and decision networks. 
$\mathbf{Linear_{R}}$ and $\mathbf{Linear_{D}}$ refer to the linear layers present in the retrieval and decision networks.

\subsubsection{Retrieval Network}
\label{rn}
The retrieval network employs an attention mechanism ~\citep{mnih2014, vaswani2017} that undergoes end-to-end optimization during training. We employ $\mathbf{CNN}$ to extract features of the current state and the states stored in the episodic memory:
\begin{equation}
S_t^{env} \mathbf{=} \mathbf{CNN}(s_t^{env})
\label{S_t^{env}}
\end{equation}
\begin{equation}
S_i^{mem} \mathbf{=} \mathbf{CNN}(s_i^{mem}), i=1,...,n
\end{equation}
Here, $s_t^{env}$ denotes the current state provided by the environment, while $s_i^{mem}$ represents a state stored in the episodic memory. $n$ signifies the length of the memory table. Subsequently, the retriever network is utilized to generate the query ($q$) and keys ($k_{1,...,n}$):
\begin{equation}
q,\, k_{1,...,n} \mathbf{=} \mathbf{Retriever}(S_t^{env}, \, S_{1,...,n}^{mem})
\end{equation}
Following this, attention scores are computed for each state within the episodic memory.
These scores gauge the relevance between the current state and the states stored in the episodic memory.
\begin{equation}
s_i \mathbf{=} q^Tk_i, i=1,...,n
\end{equation}
\begin{equation}
attn \mathbf{=} \mathbf{Softmax}(s)
\end{equation} 
\begin{equation}
    \bm{V_{mem}} \mathbf{=} attn^T v^{mem}
\end{equation} 
In the equations above, $v^{mem}$ represents the MC returns stored in the episodic memory, and $\bm{V_{mem}}$ is a value computed from the MC returns of relevant states in the episodic memory. The retrieval process is illustrated in the Algorithm \ref{algo_retrieval}  within Appendix \ref{algo_sec}.

\subsection{Dynamic Modification Mechanism}
\label{m_dmm} 
In this section, we offer a thorough explanation of the utilization of $\bm{V_{mem}}$ during the training process and the dynamic modification mechanism to foster collaboration between the parameterized retrieval network and the decision network.
The utilization of $\bm{V_{mem}}$ obtained via the retrieval network is confined to the training phase. 
During inference, only the actor network is active for decision-making.
In this section, we offer a comprehensive explanation of how the integration of $\bm{V_{mem}}$ takes place during the training phase.
For the sake of clarity, we will represent the actor and critic networks as $\bm{\pi_{\theta}}$ and $\mathbf{V}$, respectively.
It is noteworthy that our complete training process is end-to-end; gradients from $\bm{\pi_{\theta}}$ and $\mathbf{V}$ are back-propagated to $\mathbf{Hypernet}$ and $\mathbf{CNN}$, leading to the update of their respective parameters.
The training procedure of EMPPOhypernet is outlined in the Algorithm \ref{algo_method}  within Appendix \ref{algo_sec}.

To begin, $S_t^{env}$ obtained in Equation \ref{S_t^{env}} is fed into the decision network obtained in Equation \ref{Retriever_w} \ref{Retriever_R} \ref{Decision}, thereby incorporating task-specific information into $S_t^{env}$. 
Subsequently, it passes through both $\bm{\pi_{\theta}}$ and $\mathbf{V}$ to produce $a_t$ and $V_t$:
\begin{equation}
    \label{d_t}
    d_t \mathbf{=} \mathbf{Decision}(S_t^{env})
\end{equation}
\begin{equation}
\label{a_t}
a_t \mathbf{\sim} \bm{\pi_{\theta}}(a|d_t)
\end{equation}
\begin{equation}
\label{V_t}
V_t \mathbf{=} \mathbf{V}(d_t)
\end{equation}
Subsequently, we offer details for updating $\bm{\pi_{\theta}}$ and $\mathbf{V}$. Regarding $\mathbf{V}$, we employ $\bm{V_{mem}}$ as an auxiliary estimate, similar to prior works ~\citep{lin2018, kuznetsov2021}. 
Notably, our approach differs by not considering $\alpha$ as a hyperparameter; instead, we regard it as a dynamic parameter influenced by the attention scores assigned to states within the episodic memory.
If the threshold $\mathit{thr}$ is surpassed, we magnify the modifying impact of $\bm{V_{mem}}$ on $\mathbf{V}$.
The expression $\alpha(V_t - \bm{V_{mem}})^2$ signifies our modification to the original $\mathbf{V}$ loss. Subsequently, we update $\mathbf{V}$  employing this modified loss:
\begin{equation}
\fontsize{9.5}{12}\selectfont
\mathcal{L}(\mathbf{V})\mathbf{=} (1-\alpha)\delta_t^2 + \alpha(V_t-\bm{V_{mem}})^2
\label{loss}
\end{equation}
\begin{equation}
\label{delta}
\delta_t = r_t + \gamma V_{t+1} - V_t
\end{equation}
\begin{equation}
\fontsize{9.5}{12}\selectfont
\alpha \mathbf{=} \mathbf{Max}(\mathbf{Max}(attn_1,...,attn_n), \mathit{thr})
\end{equation}
Here, $r_t$ represents the immediate reward, $\gamma$ denotes the discount factor, $V_{t+1}$ and $V_t$ denotes the value computed in Equation \ref{V_t} of the succeeding and current states.
Utilizing dynamic $\alpha$ can enhance the impact of important information on the agent when the information provided is highly relevant to the current state, leading to faster network convergence. 
Our ablation experiments demonstrate that \textbf{\textit{Dynamic Modification Mechanism}} proves beneficial. About $\bm{\pi_{\theta}}$, we employ the Generalized Advantage Estimation (GAE) to compute the policy gradient of $\bm{\pi_{\theta}}$, as introduced by ~\citep{schulman2015gae}. A detailed description is provided in the appendix \ref{policy}.

\section{Experiment}

\subsection{Experimental Setting}

We evaluate our method on the MiniGrid environment ~\citep{minigrid} due to its provision of natural language instructions, which enhance the agent's grasp of the task.
Moreover, the environment includes numerous tasks that allow for knowledge-sharing among them. 
Minigrid offers researchers uncomplicated and adaptable grid world environments to conduct Reinforcement Learning research. 
These 2D environments comprise goal-oriented tasks, with vocabulary instruction to the agent. 
The tasks vary from solving different maze maps to interacting with diverse objects, including doors, keys, or boxes.

We ensure consistency in network architectures for all networks used in our experiments. 
A single 3090 NVIDIA card is used to run all experiments. 
Every environment runs for 200,000 frames. 
Evaluation is performed every 2,000 frames, with an average of 16 episodes using different random seeds.
we report the mean and 1/2 std of 16 seeds.
We plot the training curve to monitor convergence trends and focus on the average reward gathered in the last 10,000 frames.

\subsection{Evaluation Results}

Our agent model is trained on multitask learning scenarios that involve MT3 (LavaGap, RedBlueDoors, Memory), MT5 (DoorKey, LavaGap, RedBlueDoors, DistShift, Memory), and MT7 (DoorKey, DistShift, RedBlueDoors, LavaGap, Memory, SimpleCrossing, MultiRoom); The episodic memory retains data from all these tasks. In the appendix \ref{desc}, we present the language descriptions of these tasks. 
We compare our algorithm with four baselines:

(1) \textbf{PPO}: This approach utilizes PPO ~\citep{schulman2017} to train the agent model to learn multiple tasks simultaneously.\\
(2) \textbf{PPOHypernet}: This approach employs a hypernetwork to support the decision network ~\citep{beck2023}. Notably, PPOHypernet is developed within the actor-critic framework, distinguishing it from the policy-based framework of ~\citep{beck2023}.\\
(3) \textbf{EMPPO-para}: We implement the EMAC ~\citep{kuznetsov2021} within the PPO framework. However, in contrast to the original retrieval process of EMAC using random projection, we utilize a parameterized retriever. Notably, EMPPO's retriever concatenates the query and keys with the task vector.\\
(4) \textbf{EMPPO-moe}: Building upon EMPPO, we introduce a mixture-of-expert (moe) mechanism ~\citep{fedus2022review}. EMPPO-moe consists of three experts, each equipped with an identical retriever, alongside a learned gating network responsible for aggregating their outputs. The gating network takes the task vector as input. 

Table \ref{table_1} shows the average reward of the last 10,000 frames, and the training curves of EMPPO, PPOHypernet, EMPPO-para, EMPPO-moe, and EMPPOHypernet(ours) are shown in Figure \ref{curve_1} within Appendix \ref{curve}. 
Our experimental results demonstrate that EMPPOHypernet significantly enhances both the sampling efficiency and PPO performance. 
Notably, our method achieves significantly higher episode reward than our best baseline, converging faster than the baselines. Importantly, these advantages become more pronounced as the number of tasks increases.
\begin{table*}[!htbp]
\centering
\begin{tabular}{cccccc}
\toprule[\heavyrulewidth] 
Scenario & PPO & PPOHypernet & EMPPO-para & EMPPO-moe & EMPPOHypernet \\
\midrule[\lightrulewidth]
MT3 & $0.41 \pm 0.12$ & $0.47 \pm 0.09$ & $0.39 \pm 0.12$ & $\textbf{0.51} \pm 0.11$ & $0.48 \pm 0.12$  \\
MT5 & $0.12 \pm 0.06$ & $0.45 \pm 0.10$ & $0.38 \pm 0.10$ & $0.51 \pm 0.12$ & $\textbf{0.61} \pm 0.10$  \\
MT7 & $0.06 \pm 0.05$ & $0.17 \pm 0.08$ & $0.19 \pm 0.08$ & $0.27 \pm 0.09$ & $\textbf{0.60} \pm 0.09$  \\
\bottomrule[\heavyrulewidth]
\end{tabular}
\caption{PPO, PPOHypernet, EMPPO-para, EMPPO-moe, and EMPPOHypernet(ours)'s average reward of the last 10,000 frames over 16 episodes from different seeds. $\pm$ corresponds to 1/2 std over 16 episodes from different seeds.}
\label{table_1}
\end{table*}

\subsection{Ablation and Analysis}

To better understand the distinct advantages provided by each component in EMPPOHypernet, we conduct ablation studies to analyze each component's benefits. 

\subsubsection{Task-Conditioned Hypernetwork}

In EMPPOHypernet, we utilize a task-conditioned hypernetwork to augment the parameter sensitivity of the retrieval network across diverse tasks. 
Here, we explore the impacts of incorporating the task-conditioned hypernetwork by using three distinct strategies for assigning parameters to $\mathbf{Linear_{R}}$ in $\mathbf{Retriever} \mathbf{=} \mathbf{Linear_{R}} \rightarrow \mathbf{ReLU}$.

(1) \textbf{EMPPO-random}: Parameters of the retriever are randomly assigned.\\
(2) \textbf{EMPPO-para}: The retriever's parameters are updated during training.\\
(3) \textbf{EMPPOHypernet}: The parameterized retriever receives its parameters from a hypernetwork, which undergoes update during training.

Table \ref{table_3} shows the average reward of the last 10,000 frames, and the training curves of EMPPO-random, EMPPO-para, EMPPOHypernet are shown in Figure \ref{curve_3} within Appendix \ref{curve}.\\
\textbf{Conclusion}: The results indicate that EMPPOHypernet outperforms EMPPO-para, which in turn outperforms EMPPO-random. 
This underscores the advantages of updating the retriever during training. 
Moreover, the considerable enhancement achieved by the hypernetwork in terms of the retrieval network's parameter sensitivity across diverse tasks highlights its capability to direct the retriever's focus toward the unique aspects of individual tasks.  
\begin{table}[!htbp]
\centering
\begin{tabular}{cccc}
\toprule[\heavyrulewidth]
Scenario & EMPPO-random  & EMPPO-para & EMPPOHypernet \\
\bottomrule[\heavyrulewidth]
MT3  & $0.42 \pm 0.12$ &  $0.39 \pm 0.11$ & $\textbf{0.48} \pm 0.12$  \\
MT5  & $0.32 \pm 0.11$ & $0.38 \pm 0.11$ & $\textbf{0.61} \pm 0.10$  \\
MT7  & $0.03 \pm 0.03$ & $0.19 \pm 0.08$ & $\textbf{0.60} \pm 0.09$  \\
\bottomrule[\heavyrulewidth]
\end{tabular}
\vspace{7pt}
\caption{EMPPO-random, EMPPO-para, and EMPPOHypernet 's average episode reward of the last 10,000 frames over 16 episodes from different seeds. $\pm$ corresponds to 1/2 std over 16 episodes from different seeds.}
\label{table_3}
\end{table}

\vspace{-0.3cm}
\subsubsection{Dynamic Modification Mechanism}
\label{section:dmm}

The pursuit of optimal coordination between the retrieval network and the decision network necessitates the implementation of a dynamic modification mechanism within our methodology. 
The loss function of the critic network is defined as $\mathcal{L}(\mathbf{V}) \mathbf{=} (1-\alpha)\delta_t^2 + \alpha(V_t-\bm{V_{mem}})^2$.
Here, $\alpha$ is considered a dynamic parameter that undergoes adjustments based on the relevance between the current state and the states stored in the episodic memory.
While preceding algorithms using episodic memory ~\citep{lin2018, hansen2018, kuznetsov2021, chen2022, hu2021} typically treated $\alpha$ as a fixed hyperparameter, our approach dynamically adapts it.
To evaluate the effectiveness of this dynamic modification mechanism, we select a subset of tasks from the MiniGrid environment, and agents are trained on these individual tasks. 
The variant employing a consistent $\alpha$ is designated as SEMPPO, whereas the version incorporating a dynamically varying $\alpha$ is labeled DEMPPO. 
The average reward of the last 10,000 frames is presented in Table \ref{table_4}, and the training curves of PPO, SEMPPO, and DEMPPO are shown in Figure \ref{curve_2} within Appendix \ref{curve}.\\
\textbf{Conclusion:} The incorporation of a dynamic modification mechanism proves to be advantageous in introducing episodic memory to reinforcement learning algorithms. This facilitates optimal coordination between the retrieval and decision networks and alleviates the need for hyperparameter fine-tuning.

\begin{table*}[!htbp]
\centering
\begin{tabular}{lccccc}
\toprule[\heavyrulewidth] 
Task & PPO & SEMPPO ($\alpha$=0.1) & SEMPPO ($\alpha$=0.2) & DEMPPO \\
\toprule[\heavyrulewidth]
DoorKey & $0.71 \pm 0.05$  & $0.91 \pm 0.01$ & $0.92 \pm 0.00$ & $\textbf{0.93} \pm 0.00$ \\
LavaGap & $0.86 \pm 0.03$ & $0.80 \pm 0.06$ & $0.78 \pm 0.07$ & $\textbf{0.90} \pm 0.02$ \\
Distshift & $0.91 \pm 0.01$ & $0.91 \pm 0.01$ & $0.91 \pm 0.01$ & $0.91 \pm 0.01$ \\
Memory & $0.48 \pm 0.12$ & $0.50 \pm 0.12$ & $0.49 \pm 0.12$ & $\textbf{0.51} \pm 0.12$ \\
RedBlueDoor & $0.06 \pm 0.04$ & $0.14 \pm 0.07$ & $0.23 \pm 0.08$ & $\textbf{0.34} \pm 0.09$ \\
Unlock & $0.36 \pm 0.08$ & $0.92 \pm 0.01$ & $0.88 \pm 0.02$ & $\textbf{0.93} \pm 0.01$ \\
\bottomrule[\heavyrulewidth]
\end{tabular}
\caption{PPO, SEMPPO, and DEMPPO 's average episode reward of the last 10,000 frames over 16 episodes from different seeds. $\pm$ corresponds to 1/2 std over 16 episodes from different seeds.}
\label{table_4}
\end{table*}

\section{Conclusion}
In this work, we develop Actor-Critic with Multitask Episodic Memory Based on Hypernetwork, an algorithm incorporating a hippocampus module to enhance agents' decision-making. According to the varying activation features displayed by the hippocampus in response to different tasks~\citep{davachi2003}, we utilize a task-conditioned hypernetwork to adapt the retrieval network to ensure that the retrieved information is most relevant to the specific task. Moreover, we propose a dynamic modification mechanism to facilitate collaboration between the retrieval and decision networks. Our results demonstrate that including the hippocampus-like structure in our algorithm significantly enhances the agent's performance. Ablations demonstrate the effectiveness of each component.


\clearpage
\appendix

\section{Algorithm Pseudo-code}
\label{algo_sec}

\begin{algorithm*}[ht]
    \caption{Actor-Critic with Multitask Episodic Memory Based on Hypernetwork(training procedure)}
    \label{algo_method}
    \begin{algorithmic}
    \STATE \textbf{\textrm{Step 1:}} Initialize $\bm{\pi_{\theta}}$, $\mathbf{V}$, $\mathbf{Hypernet}$ and $\mathbf{CNN}$.
    \vspace{3pt} 
    
    \STATE \textbf{\textrm{Step 2:}} Initialize replay buffer $B_R$, replay buffer $B$, and episodic memory $M$.
    \vspace{3pt} 
    
    \STATE \textbf{\textrm{Step 3:}} Collect data and update parameters of $\bm{\pi_{\theta}}$, $\mathbf{V}$, $\mathbf{Hypernet}$ and $\mathbf{CNN}$.
        \FOR{$epoch \mathbf{=} 1$ to $N$}
            \STATE \textbf{\textrm{Step 3-1}}: Interact with the environment using the current $\bm{\pi_{\theta}}$ to collect data.
                \FOR{$t \mathbf{=} 1$ to $\mathrm{step}$}
                    
                    \vspace{3pt}
                    \STATE \textbf{\textrm{Step 3-1-1}}: Interact with the environment.
                    \STATE Receive the current state $s_{t}^{env}$, task $T$ and extract features of $s_{t}^{env}$ to get $S_{t}^{env}$(Equation \ref{S_t^{env}}).
                    \STATE Generate parameters for $\mathbf{Decision}$ network(Equation \ref{Retriever_w} \ref{Retriever_R} \ref{Decision}).
                    \STATE Select action $a_t$ and get value of the current state $V_t$(Equation \ref{d_t} \ref{a_t} \ref{V_t}).
                    \STATE Receive next state $s_{t+1}^{env}$, reward $r_t$, done $d_t$.
                    \STATE Store $(s_{t}^{env}, T, a_t, V_t, r_t, s_{t+1}^{env}, d_t)$ in $B$.
                    
                    \vspace{3pt}
                    \STATE \textbf{\textrm{Step 3-1-2}}: Populate data into replay buffer $B_R$ and episodic memory $M$.
                    \IF{$d \mathbf{=}\mathbf{=} True \, \OR \, t\mathbf{=}\mathbf{=} step$}
                        \STATE Initialize $R = 0$.
                        \STATE Reverse temporary buffer $B$. 
                        \FOR{$i = 1$ to $\mathbf{len}(B)$}
                            \STATE Take out $(s_i, T_i, a_i, V_i, r_i, s_{i+1})$ from $B$.
                            \STATE Compute discounted episode return $ R_i \mathbf{=} R*\gamma + r_i$ and update $ R \mathbf{=} R_i$.
                            \STATE Compute $A_i^{GAE}$(Equation \ref{delta} \ref{A_t}).
                            \STATE Store $(s_i, T_i, a_i, r_i, s_{i+1}, A_i^{GAE})$ in $B_R$ and $(s_i, T_i, R_i)$ in $M$.
                        \ENDFOR
                        \STATE Free $B$.
                    \ENDIF
                    
                \ENDFOR
            \vspace{3pt}   
            
            \STATE  \textbf{\textrm{Step 3-2}}: Update parameters of $\bm{\pi_{\theta}}$, $\mathbf{V}$, $\mathbf{Hypernet}$ and $\mathbf{CNN}$ using data collected.
                \FOR{$i = 1$ to $\mathbf{len}(B_R)$}
                    \STATE Take out $(s_i, T_i, a_i, r_i, s_{i+1}, A_i^{GAE})$ from $B_R$.
                    \STATE Compute $\bm{V_{mem}}$ of current state according to Algorithm \ref{algo_retrieval}.
                    \STATE Compute loss for $\mathbf{V}$(equation \ref{loss}) and policy gradient for $\bm{\pi_{\theta}}$(equation \ref{J}).
                    \STATE Perform gradient backpropagation and update parameters of $\bm{\pi_{\theta}}$, $\mathbf{V}$, $\mathbf{Hypernet}$ and $\mathbf{CNN}$.
                \ENDFOR
            \vspace{3pt}
            
            \STATE  \textbf{\textrm{Step 3-3}}: If $B_R$ and $M$ reach full capacity, purge some older data.
        \ENDFOR
  \end{algorithmic}
\end{algorithm*}

\begin{algorithm}
  \caption{Compute $\bm{V_{mem}}$ utilizing the retrieval network}
  \label{algo_retrieval}
  \begin{algorithmic}
            \STATE Generate parameters for $\mathbf{Retriever}$(Equation \ref{Retriever_w} \ref{Retriever_R} \ref{Retriever}).
            \STATE  Compute $\bm{V_{mem}}$:

                \STATE \quad \textbf{\textrm{Step 1:}} Extract features of $s_t^{env}$ and $s_{1,...,n}^{mem}$.
                \STATE \quad  $S_t^{env} \mathbf{=} \mathbf{CNN}(s_t^{env})$
                \STATE \quad  $S_i^{mem} \mathbf{=} \mathbf{CNN}(s_i^{mem}), i=1,...,n$
                \vspace{3pt}
            
                \STATE \quad \textbf{\textrm{Step 2:}}  $q$ and $k_{1,...,n}$.
                \STATE \quad  q,  $k_{1,...,n} \mathbf{=} \mathbf{Retriever}(S_t^{env}, \, S_{1,...,n}^{mem})$
                \vspace{3pt}
                
                \STATE \quad \textbf{\textrm{Step 3:}}  $attn$ between $q$ and $k_{1,...,l_M}$.
                \STATE \quad $s_i \mathbf{=} q^T k_i, i=1,...,n$
                \STATE \quad  $attn \mathbf{=} \mathbf{Softmax}(s)$
                \vspace{3pt}
                
                \STATE \quad \textbf{\textrm{Step 4:}}  $\bm{V_{mem}}$.
                \STATE \quad  $\bm{V_{mem}} \mathbf{=} attn^T v^{mem}$             
  \end{algorithmic}
\end{algorithm}
\vspace{5cm}

\section{Training of Policy}
\label{policy}
About $\bm{\pi_{\theta}}$, we employ the Generalized Advantage Estimation (GAE) to compute the policy gradient of $\bm{\pi_{\theta}}$, as introduced by ~\citep{schulman2015gae}. GAE is computed as follows:
\begin{equation}
\label{A_t}
A_t^{GAE} \mathbf{=} \delta_t + (\gamma \lambda) \delta_{t+1} + \cdots + (\gamma \lambda)^{T-t+1} \delta_{T-1}
\end{equation}
Here, $\gamma$ denotes the discount factor, $\lambda$ governs the impact of future rewards on the advantage calculation, and $\delta_{t,...,T-1}$ is computed in Equation \ref{delta}.
The policy gradient is computed as follows:
\begin{equation}
\label{J}
\nabla J(\theta) \mathbf{=} \mathbb{E} \left[ \frac{\bm{\pi_{\theta}}(a|s)}{\bm{\pi_{\theta}^{old}}(a|s)} \cdot A_t^{GAE} \cdot \nabla \log \bm{\pi_{\theta}}(a|s) \right]
\end{equation}
Here, $\theta$ is the parameter of $\bm{\pi_{\theta}}$, $\bm{\pi_{\theta}}(a|s)$ represents the probability of selecting action based on the current policy, $\bm{\pi_{\theta}^{old}}(a|s)$ is the corresponding probability according to the old policy. The computed policy gradient is then employed to update the policy parameters $\theta$.

\section{Minigrid Task Descriptions}
\label{desc}
\begin{figure*}[!ht]
\centering 
\includegraphics[width=1.00\linewidth]{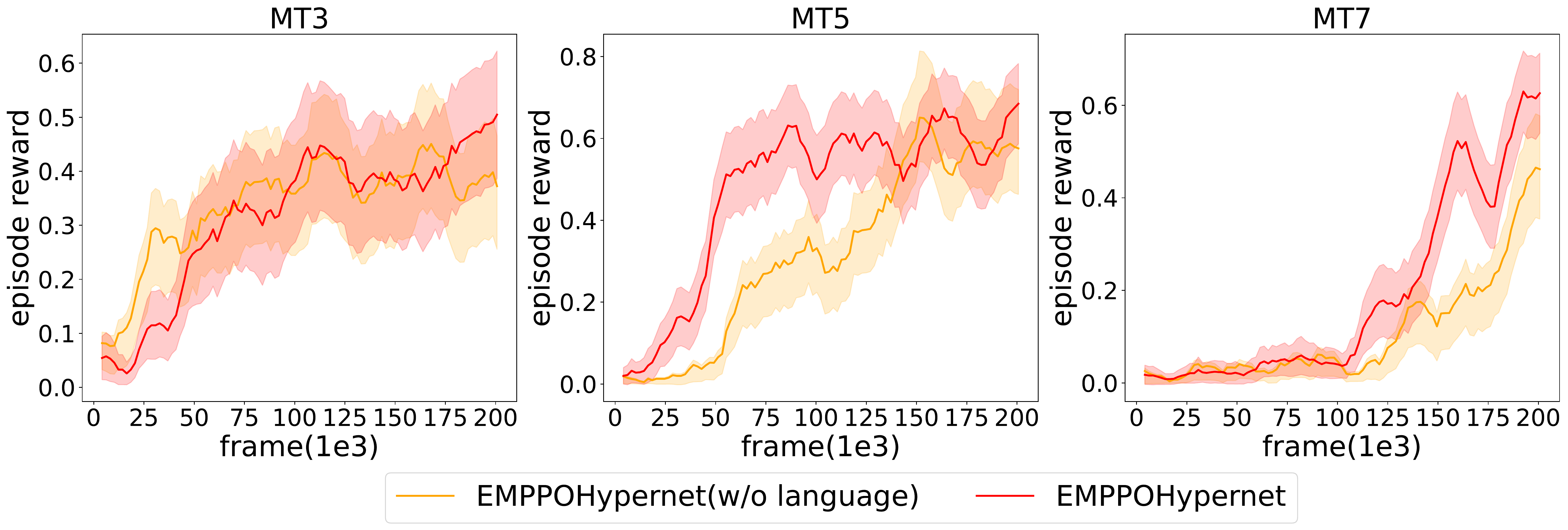}
\caption{EMPPOHypernet(w/o language), and EMPPOHypernet's learning curve.}
\label{curve_4}
\end{figure*}

Language descriptions of minigird tasks in our experiments are detailed in table \ref{task detail table}.
We conduct an ablation experiment on the introduction of language.
We provide two versions of our algorithm for comparison: one with natural language guidance and another without such guidance to demonstrate the benefits of introducing natural language guidance. 
Table \ref{table_lang} shows the average reward of the last 10,000 frames the training curves of  are shown in Figure \ref{curve_4}.
Our experimental results demonstrate that natural language guidance supports agents in better understanding the task, which leads to enhanced performance.

\begin{table}[!htbp]
\centering
\begin{tabular}{cc}
\toprule
task & Description \\
\midrule
DoorKey &  Use the key to open the door and then get to the goal. \\
DistShift &  Get to the green goal square. \\
RedBlueDoors &  Open the red door and then the blue door. \\
LavaGap &  Avoid the lava and get to the green goal square. \\
Memory &  Go to the matching object at the end of the hallway. \\
SimpleCrossing &  Find the opening and get to the green goal square. \\
MultiRoom &  Traverse the rooms to get to the goal. \\
\bottomrule
\end{tabular}
\vspace{5pt}
\caption{Minigrid Task Descriptions.}
\label{task detail table}
\end{table}
\vspace{-0.2cm}
\begin{table}[!htbp]
\centering
\begin{tabular}{p{1.5cm}cc}
\toprule[\heavyrulewidth] 
Scenario & EMPPOHypernet(w/o language) & EMPPOHypernet \\
\midrule[\lightrulewidth]
MT3 & $0.37 \pm 0.12$ & $0.48 \pm 0.12$   \\
MT5 & $0.57 \pm 0.11$ & $0.61 \pm 0.10$   \\
MT7 & $0.40 \pm 0.10$ & $0.60 \pm 0.09$   \\
\bottomrule[\heavyrulewidth]
\end{tabular}
\vspace{5pt}
\caption{EMPPOHypernet(w/o language), and EMPPOHypernet's average reward of the last 10,000 frames over 16 episodes from different seeds. $\pm$ corresponds to 1/2 std over 16 episodes from different seeds.}
\label{table_lang}
\end{table}

\section{Training Curve}
\label{curve}

\begin{figure}[H]
\centering 
\includegraphics[width=1.00\linewidth]{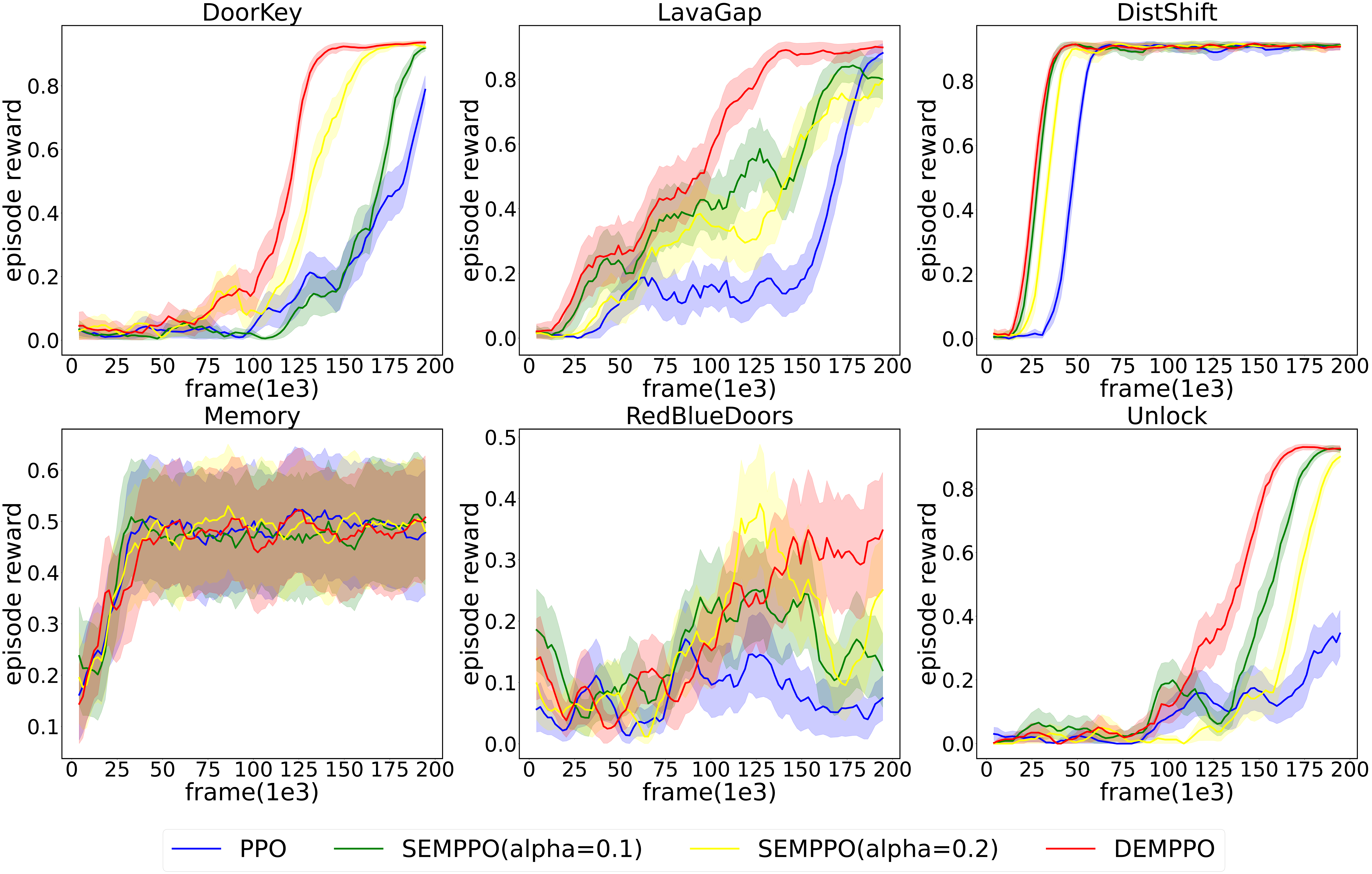}
\caption{PPO, SEMPPO, and DEMPPO's learning curve.}
\label{curve_2}
\end{figure}

\begin{figure*}[!ht]
\centering 
\includegraphics[width=1.00\linewidth]{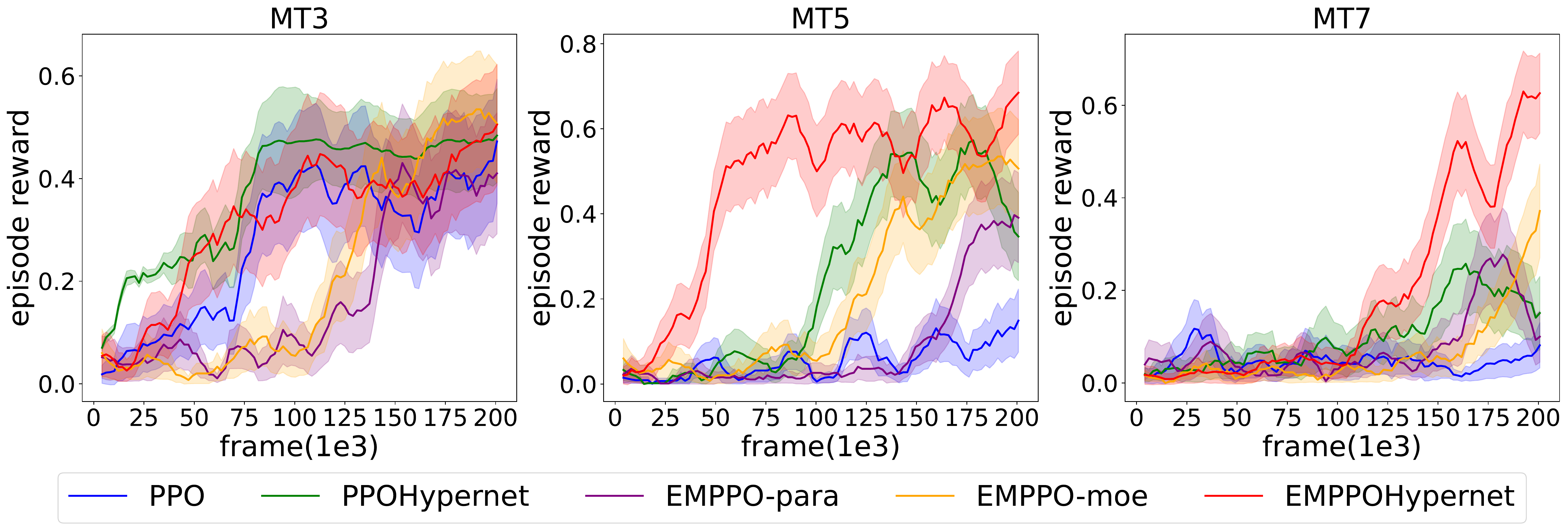}
\caption{PPO, PPOHypernet, EMPPO-para, EMPPO-moe and EMPPOHypernet(ours)'s learning curve.} 
\label{curve_1}
\end{figure*}

\begin{figure*}[!ht]
\centering 
\includegraphics[width=1.00\linewidth]{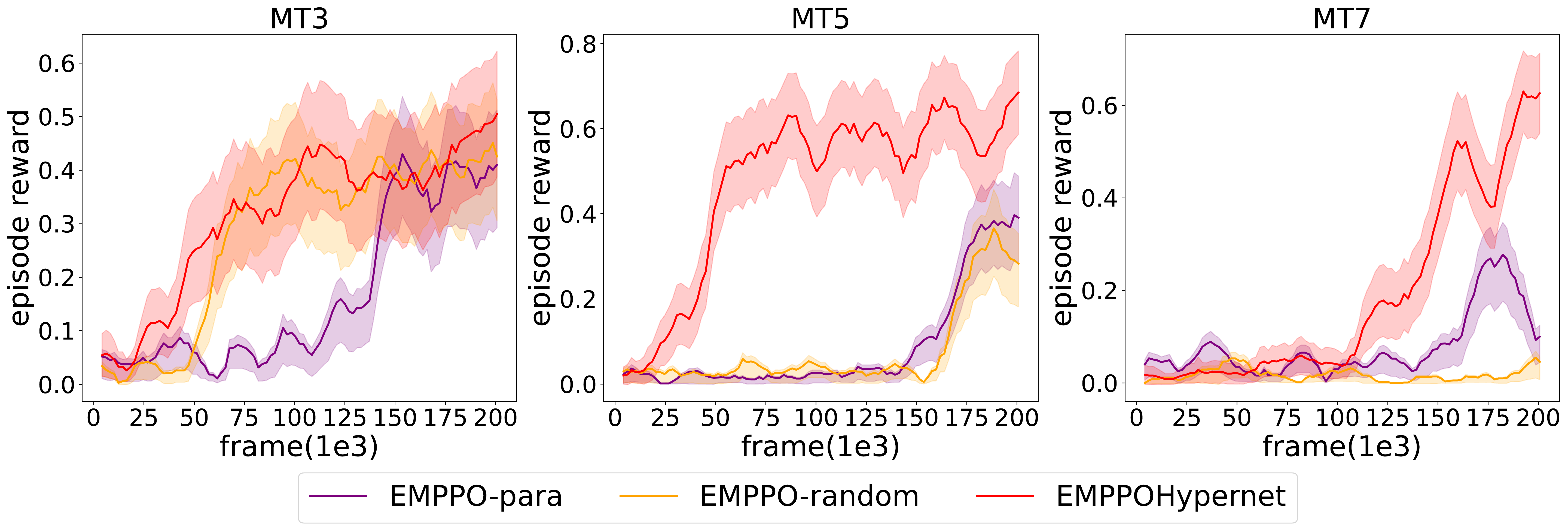}
\caption{PPO, PPOHypernet, EMPPO-para, EMPPO-moe and EMPPOHypernet(ours)'s learning curve.} 
\label{curve_3}
\end{figure*}

\section{Experimental Hyperparameters}
\begin{table}[!htbp]
\centering
\begin{tabular}{cc}
\toprule
Hyperparameters & Value \\
\midrule
Batch size for PPO & 256 \\
Discount factor & 0.99 \\
Learning rate & 0.001 \\
Lambda coefficient in GAE formula & 0.95 \\
Entropy term coefficient & 0.01 \\
Value loss term coefficient & 0.5 \\
Maximum norm of gradient & 0.5 \\
Adam and RMSprop optimizer epsilon & 1e-8 \\
RMSprop optimizer alpha & 0.99 \\
Clipping epsilon for PPO & 0.2 \\
Dimension of $T_{lang}$ & 64 \\
Dimension of $T_{id}$ & 64 \\
Size of episodic memory & 50 \\
\bottomrule
\end{tabular}
\vspace{5pt}
\caption{Hyperparameters used in our experiments.}
\label{table_hyperparameter}
\end{table}

\vspace{20pt}
\section{Notation}
\label{table_notation}
\begin{table*}[!htbp]
\centering
\begin{tabular}{cc}
\toprule
Notation & Description \\
\midrule
EMDQN &  Episodic Memory Deep Q-network\\
EMAC &   Episodic Memory Actor-Critic \\
EMPPO-random & Parameters of the retriever are randomly assigned \\
EMPPO-pare &  The retriever's parameters are updated during training \\
EMPPO-moe &  Building upon EMPPO, moe mechanism is introduced \\
moe & mixture-of-expert \\
MC &   Monte Carlo \\
MFEC &  Model Free Episodic Control \\ 
NEC &   Neural Episodic Control \\
PPOHypernet & A variant of our method that excludes retrieval network \\
SEMPPO &  Fixed $\alpha$ in Equation 15\\
DEMPPO &  Dynamic $\alpha$ in Equation 15 \\

$\bm{\pi_{\theta}}$ & Actor network \\
$\mathbf{V}$ & Critic network \\

$n$ &  Length of the memory table \\
$s_i^{mem}$ &  State in the episodic memory \\
$S_{1,...,n}^{mem}$ &  $s_{1,...,n}^{mem}$'s feature extracted by $\mathbf{CNN}$\\
$s_t^{env}$ &  State provided by the environment \\
$S_t^{env}$ &  $s_t^{env}$'s feature extracted by $\mathbf{CNN}$\\
$t_i^{mem}$ &  Task in the episodic memory \\
$v_i^{mem}$ &  MC returns in the episodic memory \\
$\bm{V_{mem}}$ & Estimate value calculated by Algorithm 1 \\

$attn$ & $s$ are normalized through softmax.\\ 
$t_{id}$, $t_{lang}$ & Task ID and language description \\

${W_{R}}{W_{D}}$ & Matrices of $\mathbf{Retriever}$ and $\mathbf{Decision}$ \\
${b_{R}}, {b_{D}}$ & Biases of $\mathbf{Retriever}$ and $\mathbf{Decision}$ \\

$q, k_{1,...,n}$ & Input $S_t^{env}$ and $S_{1,...,n}^{mem}$ into $\mathbf{Retriever}$ to obtain\\  
$s$ & Dot product of $q$ and $k_{1,...,n}$ \\
$attn$ & $s$ are normalized through softmax.\\

$\mathbf{\pi_{\theta}}$, $\mathbf{V}$ & The actor and critic networks \\
$d_t$ & Input $S_t^{env}$ into $\mathbf{Decision}$ to obtain \\
$a_t$, $V_t$ &  Input $d_t$ into $\mathbf{\pi_{\theta}}$ and $\mathbf{V}$ to obtain\\
$\mathcal{L}(\mathbf{V})$ & Loss function of $\mathbf{V}$ \\
$\delta_t$ & $\delta_t = r_t + \gamma V_{t+1} - V_t$ \\
$\gamma$  & The discount factor \\
$\alpha$ & The proportion of $(V_t - V_{mem})^2$ in $\mathcal{L}(\mathbf{V})$\\
$\mathit{thr}$ & If $\alpha$ $>$ $\mathit{thr}$, we enhance $V_{mem}$'s impact on $\mathbf{V}$\\ 

$A_t^{GAE}$ &  Generalized Advantage Estimation \\
$\lambda$ & The impact of future rewards on the advantage calculation\\
$\nabla J(\theta)$ & The policy gradient of $\mathbf{\pi_{\theta}}$\\

$B_R$ & Replay buffer\\
$B$ & Replay buffer\\
$M$ & Episodic memory \\

\bottomrule
\end{tabular}
\vspace{5pt}
\caption{Summary of notations.}
\end{table*}

\end{document}